\documentclass[letterpaper, 10 pt, conference]{ieeeconf}  

\IEEEoverridecommandlockouts                              

\usepackage{caption} 
\usepackage{graphics} 
\usepackage{epsfig} 
\usepackage{mathptmx} 
\usepackage{times} 
\usepackage{amsmath} 
\usepackage{amssymb}  
\usepackage[ruled,vlined,linesnumbered]{algorithm2e}
\usepackage{wrapfig} 
\usepackage{multirow}   
\usepackage{array}      
\usepackage{color}

\title{\LARGE \bf
A 4D Radar Camera Extrinsic Calibration Tool Based on 3D Uncertainty Perspective N Points.
}

\author{Chuan Cao, Xiaoning Wang, Wenqian Xi, Han Zhang$^{*}$, Weidong Chen and Jingchuan Wang
\thanks{Chuan Cao, Han Zhang, Weidong Chen and Jingchuan Wang are with School of Automation and Intelligent Sensing, Institute of Medical Robotics, Shanghai Jiao Tong University, Shanghai 200240, China Key Laboratory of System Control and Information Processing, Ministry of Education of China, Shanghai 200240. Xiaoning Wang is with Ruijin Hospital, Shanghai Jiao Tong University School of Medicine, China. Wenqian Xi is with Renji hospital, Shanghai Jiao Tong University School of Medicine, China. This work is supported by Shanghai 2024 "Science and Technology Innovation Action Plan" Special Project on Elderly Care Technology Support:  24YL1900800. *Corresponding Author: $zhanghan\_tc@sjtu.edu.cn$}%
}

\begin{document}

\maketitle
\thispagestyle{empty}
\pagestyle{empty}

\begin{abstract}

4D imaging radar is a type of low-cost millimeter-wave radar(costing merely 10-20$\%$ of lidar systems) capable of providing range, azimuth, elevation, and Doppler velocity information. Accurate extrinsic calibration between millimeter-wave radar and camera systems is critical for robust multimodal perception in robotics, yet remains challenging due to inherent sensor noise characteristics and complex error propagation. This paper presents a systematic calibration framework to address critical challenges through a spatial 3d uncertainty-aware PnP algorithm (3DUPnP) that explicitly models spherical coordinate noise propagation in radar measurements, then compensating for non-zero error expectations during coordinate transformations. Finally, experimental validation demonstrates significant performance improvements over state-of-the-art CPnP baseline, including improved consistency in simulations and enhanced precision in physical experiments. This study provides a robust calibration solution for robotic systems equipped with millimeter-wave radar and cameras, tailored specifically for autonomous driving and robotic perception applications.

\end{abstract}

\section{INTRODUCTION}
Accurate extrinsic calibration between 4D millimeter-wave radar and camera systems is critical for robust multimodal perception in safety-critical robotics\cite{b9}\cite{b24}. Robotic systems with sufficient safety redundancy typically integrate heterogeneous sensors and multimodal measurement data into their perception frameworks. This necessitates precise spatio alignment between sensors, particularly requiring accurate determination of the spatial transformation parameters (6 degrees of freedom, 6-DoF)\cite{b7}. While factory calibration provides initial parameters, inevitable mechanical wear and maintenance interventions during operation necessitate recalibration. Hence, developing an accurate, reliable, and low-cost calibration methodology becomes imperative. This paper presents a multi-sensor calibration framework specifically addressing extrinsic parameter estimation between 4D millimeter-wave radar and camera systems.

Cameras provide rich geometric and textural features with inherent spatial relationships through photometric projection onto image planes. However, their passive operation principle and scale ambiguity induced by projective transformation render them susceptible to illumination variations and inadequate for spatial metric estimation.

\begin{figure}[!htpb]
\centering
\includegraphics[width=3.0in]{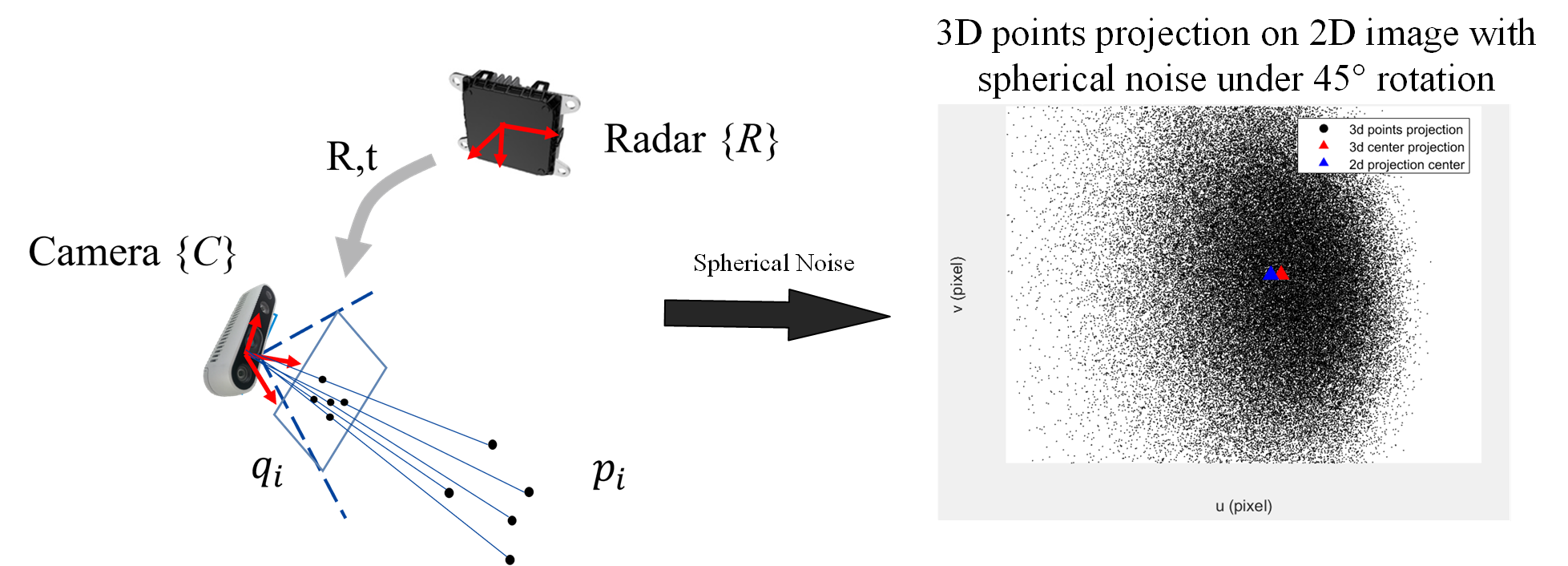}
\caption{Illustration of the PnP problem in radar-camera calibration and consistency bias refer to spherical noise.}
\label{pnp-illustration}
\end{figure}

Integrating millimeter-wave radar effectively addresses these limitations. Operating through active emission of electromagnetic radiation at millimeter wavelengths, radar demonstrates inherent robustness against adverse photometric conditions. Technological advancements, exemplified by the ZF FRGen21 4D imaging radar, enable acquisition of 4D information (3D spatial coordinates + Doppler velocity), resolving elevation angle ambiguity inherent in conventional 3D imaging radars (planar coordinates + Doppler velocity). Furthermore, radar-derived relative velocity measurements of environmental targets enhance robotic perception dimensionality. The synergistic fusion of radar and camera modalities enables complementary perception capabilities across both nominal and visually degraded conditions.

However, millimeter-wave radar integration introduces two principal challenges: 1. Reduced resolution and elevated noise: Radar measurements exhibit substantially lower spatial resolution compared to visual data, with inherent noise levels significantly higher under nominal conditions. Additionally, persistent artifacts from multipath reflections manifest as outliers resistant to conventional filtering techniques. 2. Spherical coordinate noise propagation: Radar measurements acquired in spherical coordinates undergo nonlinear coordinate transformations, inducing non-zero noise expectations that critically impact subsequent calibration accuracy(Fig 1).

The principal contributions of this work are following:
\begin{itemize}
    \item A systematic calibration framework enabling accurate, efficient, and cost-effective extrinsic calibration between millimeter-wave radar and camera systems.
    \item A spatial 3d uncertainty-aware PnP algorithm addressing measurement noise characteristics in spherical coordinate systems.
\end{itemize}

This paper is organized as follows: Section I introduces basic 4D radar camera calibration concepts. Section II reviews state-of-the-art calibration systems from two perspectives: architectural design paradigms and evolutionary trends in PnP algorithms. Section III elaborates on the proposed methodology. Section IV presents simulation and physical experimental validation. Finally, Section V concludes with research summary and future directions.

\section{RELATED WORKS}
In this section, we survey the spatio extrinsic calibration algorithms that can be applied to mm-wavelength radars in relation to cameras. Part A reviews radar camera calibration system design, and Part B introduce classical and SOTA PnP Algorithms to estimate spatio pose based on 3d and 2d pairs.

\subsection{Radar Camera Calibration System}
Calibration between radar and other sensors is necessary to maintain consistent and accurate results in multisensor fusion applications\cite{b1}, but facing some new challenges including sparse detection, and the density of detection is not sufficient to represent the physical characteristics of the target, like lines and angles\cite{b2}.

Existing calibration approaches fall into 2 systematic categories following:

\textbf{Target-based calibration methods} employ artificial markers with co-observable features. Juraj Peršić\cite{b5} designed penetrable calibration boards combining radar corner reflectors with visual patterns, subsequently, a two-step optimization approach is employed to minimize the reprojection error and radar cross-section (RCS) inconsistencies, with feasibility rigorously validated across multiple radar models. Domhof\cite{b6} also designed a calibration target featuring corner reflectors mounted behind millimeter-wave-transparent materials with perforated patterns, enabling simultaneous extrinsic calibration of radar, lidar, and cameras, achieved a joint accuracy of 0.33 degree (rotation) in controlled environments. Jun Zhang\cite{b8} addressed the Vehicle-to-Everything (V2X) road perception scenario by designing an eight-quadrant corner reflector enclosed within a foam sphere, achieving a best-case calibration accuracy of 0.07 meters under optimal conditions.

\textbf{Trajectory-based calibration methods} estimate extrinsic parameters between sensors by exploiting motion consistency derived from their synchronized trajectories. This approach eliminates the need for artificial targets, enabling extrinsic recovery through natural sensor motion patterns in dynamic environments. Emmett Wise\cite{b11} proposed a continuous B-spline trajectory modeling for 4D radar-camera systems, reducing calibration error to $<$0.5° in open-road scenarios but degrading to $>$2° in urban canyons due to multipath interference. Building upon prior work, Wise et al.\cite{b10} further integrated millimeter-wave radar ego-velocity estimates (derived from Doppler measurements) with monocular visual odometry, jointly optimizing rotation, translation, and velocity parameters to estimate extrinsic calibration. Experimental validation demonstrated accuracy comparable to target-based methods in both simulations and physical deployments. Schöller\cite{b13} proposed a calibration network for intelligent transportation systems, leveraging data-driven approaches to achieve object-level angular alignment by fusing RGB images and radar data. This framework demonstrates significant potential for enhancing multi-modal perception in dynamic traffic environments.  Peršić et al.\cite{b21} introduced a dynamic object-based calibration framework, where sensor extrinsics are estimated through track-to-track association of detected objects. The graph-based calibration is activated upon detecting rotational disturbances, ensuring adaptability in dynamic environments. In a follow-up study, Peršić et al.\cite{b12} employed Gaussian processes to jointly track targets across sensors, interpolating trajectories for temporal alignment. This is followed by on-manifold ICP optimization, achieving high-precision extrinsic calibration in multi-sensor systems.

\subsection{PnP Algorithms}
Perspective-n-Points (PnP) is a classical problem in computer vision that aims to estimate the pose of a camera based on known 3D points in space and their corresponding 2D projections in an image\cite{b16}. By selecting the coordinate system of the 3D points as the radar coordinate system, the pose of the camera in the radar coordinate system can be computed after obtaining matched features (typically the 3D coordinates of points in space and their corresponding 2D projection coordinates) using N pairs of such matched points.

PnP algorithms can be divided into two categories based on whether they account for uncertainty. Traditional PnP algorithms that do not consider uncertainty include DLT, P3P\cite{b20}, EPnP\cite{b18}, UPnP\cite{b23}, etc. These methods are computationally efficient but sensitive to noise in both 3D and 2D points, with the P3P algorithm being the most susceptible. On the other hand, PnP algorithms that incorporate measurement uncertainty include notable examples such as Direct Least Squares (DLS)\cite{b15}, Maximum Likelihood PnP (MLPnP)\cite{b3} based on noise distribution, EPnPU and DLSU\cite{b14} that account for uncertainty in 3D points, and CPnP\cite{b17} that ensures consistency with 2D noise on the image plane.

However, there is currently a lack of PnP algorithms specifically designed for the high-noise spherical coordinate system 3D point-to-image 2D point measurements in the context of millimeter-wave radar-camera systems. Additionally, the nonlinearity introduced by coordinate system transformations poses challenges to the consistency of existing PnP algorithms.

\section{METHODOLOGY}
\begin{figure*}[h]
\centering
\includegraphics[width=0.85\linewidth]{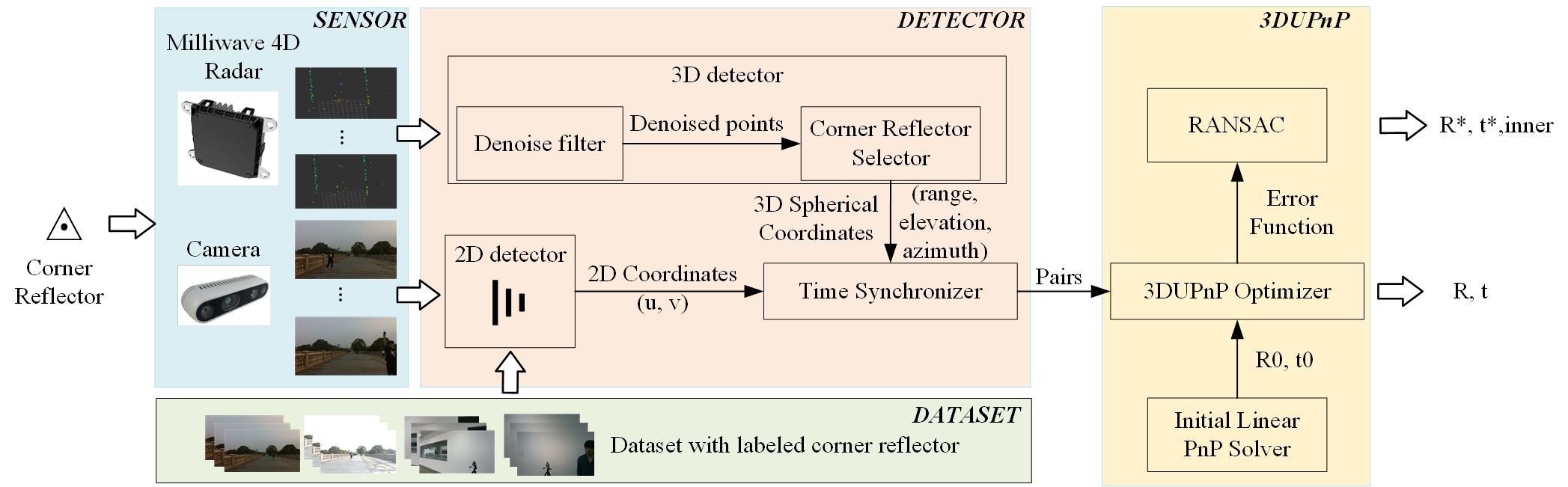}
\caption{Diagram of Millimeter-Wave Radar-Camera Calibration System.}
\label{fig:wide_figure}
\end{figure*}
As illustrated in Fig. 2, the framework detects several fixed corner reflectors, where the detection module processes synchronized images and 4D radar points to generate matched 3D-2D correspondences. These pairs are then fed into 3DUPnP for extrinsic parameter estimation, achieving robustness against both sensor noise and environmental variability.

Calibration accuracy is predominantly governed by detection and PnP errors. Compared to targetless methods, our system employs feature-rich artificial landmarks (corner reflectors) to minimize environmental dependencies and suppress detection errors. Unlike conventional PnP solvers, the following proposed 3DUPnP explicitly addresses spherical-coordinate noise propagation in 4D radar points, minimizing 3D uncertainties inherent to radar measurements.
\subsection{PnP Problem Formulation}
Suppose there are n points whose coordinates based on radar frame and setting the world coordinates on radar, $p_{i}=[x_i, y_i, z_i]^T$, $i=1,2,...n$. Denote the points' coordinates in the camera frame as $\hat{p}$: $\hat{p}_i=[\hat{x}_i, \hat{y}_i, \hat{z}_i]^T$, $i=1,2,...n$, as Fig.1 shows. Then, given the rotation
matrix R and transformation vector t from the world frame to the camera frame, it holds that $\hat{p}_i=Rp_i+t$.

Further, for a calibrated pinhole camera with the intrinsic matrix
\begin{align}
    K=\begin{bmatrix}
        f_x &0 &u_0\\
        0 &f_y &v_0\\
        0 &0 &1
    \end{bmatrix}
\end{align}
the ideal projection equation is
\begin{align}
    s_i\begin{bmatrix}
        q_i\\1
        \end{bmatrix}=K
        \begin{bmatrix}
        R &t
        \end{bmatrix}
        \begin{bmatrix}
        p_i\\1
        \end{bmatrix},
\end{align}
$s_i$ is the scale factor of i-th point, $q_i$ and $p_i$ are corresponding unbiased 2D and 3D point.\\
The standard approach minimizes the error between the observed 2D image points and the projected 3D points, typically expressed as $\| \pi(p) - q \|$. This method assumes that the noise primarily comes from the 2D image coordinates, providing consistent results as the number of correspondences increases. However, when noise originates from the 3D coordinates, scale bias from depth inaccuracies can degrade the consistency. 
To mitigate this, compensating for scale errors is necessary, improving pose estimation consistency, especially when 3D coordinate noise dominates.\\
Alexander[6] Considered the impact of noise in 3D coordinates, using the algebraic point residual 
\begin{align}
    r^{pt}(R,t,p,q)=\hat{p}^{(1:2)}(R,t,p)-q\hat{p}^{(3)}(R,t,p)
\end{align} as the optimization target. However, this method ensures consistency only if the spatial noise in the xy plane of the camera coordinate system has an expectation of zero. In the case of millimeter-wave radar and camera calibration tasks, the measurement noise of spatial points does not follow this assumption. A more general scenario is that the noise of spatial measurement points follows a Gaussian distribution in spherical coordinates.

\subsection{PnP Algorithm With Spherical Noise}
Consider 3D points represented in the spherical coordinate system $\{S\}$, where each point is parameterized by $(\rho_i, \theta_i, \phi_i)$ with $\rho_i \in \mathbb{R}^+$, $\theta_i \in [0, \pi]$, and $\phi_i \in [0, 2\pi)$. The Cartesian coordinates in radar frame $\{R\}$ relate to spherical coordinates through:

\begin{equation}
    f(r_i^R)=p_i^R = \begin{bmatrix}
        x_i^R \\ y_i^R \\ z_i^R
    \end{bmatrix} = \begin{bmatrix}
        \rho_i\sin\theta_i\cos\phi_i \\
        \rho_i\sin\theta_i\sin\phi_i \\
        \rho_i\cos\theta_i
    \end{bmatrix}
\end{equation}

When Gaussian noise $\boldsymbol{\sigma}_i = [\delta_\rho, \delta_\theta, \delta_\phi]^\top \sim \mathcal{N}(0, \boldsymbol{\Sigma}_S)$ is introduced to spherical coordinates, the perturbed coordinates measurement become:

\begin{equation}
    \Tilde{\rho}_i = \rho_i + \delta_\rho, \quad
    \Tilde{\theta}_i = \theta_i + \delta_\theta, \quad
    \Tilde{\phi}_i = \phi_i + \delta_\phi
\end{equation}
According to technical specifications provided by the radar manufacturer, we learned that the measurement noise of the millimeter-wave radar can be considered as three-dimensional independent Gaussian noise to some extent. Therefore, we can initially assume 
\begin{align}
    \boldsymbol{\Sigma}_S = diag(\epsilon_\rho, \epsilon_\theta, \epsilon_\phi)
\end{align}
Based on equation (4), 
\begin{align}
    &\mathbb{E}[\delta p_i] =\mathbb{E}[\Tilde{p}_i - p_i] \\
    &=\begin{bmatrix}
        \mathbb{E}[(\rho_i+\delta \rho_i)\sin(\theta_i+\delta \theta_i)\cos(\phi_i+\delta \phi_i)-\rho_i\sin\theta_i\cos\phi_i]\\
        \mathbb{E}[(\rho_i+\delta \rho_i)\sin(\theta_i+\delta \theta_i)\sin(\phi_i+\delta \phi_i)-\rho_i\sin\theta_i\sin\phi_i]\\
        \mathbb{E}[(\rho_i+\delta \rho_i)\cos(\theta_i+\delta \theta_i)-\rho_i\cos\theta_i]
    \end{bmatrix}\\
    &=\begin{bmatrix}
        \mathbb{E}[(\rho_i)\sin(\theta_i+\delta \theta_i)\cos(\phi_i+\delta \phi_i)-\rho_i\sin\theta_i\cos\phi_i]+0\\
        \mathbb{E}[(\rho_i)\sin(\theta_i+\delta \theta_i)\sin(\phi_i+\delta \phi_i)-\rho_i\sin\theta_i\sin\phi_i]+0\\
        \mathbb{E}[(\rho_i)\cos(\theta_i+\delta \theta_i)-\rho_i\cos\theta_i]+0
    \end{bmatrix}
\end{align}
Based on the characteristic function analysis of Gaussian noise and Cauchy's formula, the noise term $\delta_\rho$ that follows $\mathcal{N}(0,\epsilon_\rho)$ , after undergoing a cosine or sine transformation, there are $\mathbb{E}[cos(\delta_\theta)] = e^{(-\frac{\epsilon_\theta^2}{2})}$ , $\mathbb{E}[sin(\delta_\theta)] = 0$, $\mathbb{E}[\cos^2(\delta_\theta)] = \frac{1+e^{-2\epsilon_\theta^2}}{2}$, $\mathbb{E}[\sin^2(\delta_\theta)] = \frac{1-e^{-2\epsilon_\theta^2}}{2}$, $\mathbb{E}[\sin(\delta_\theta)\cos(\delta_\theta)] = 0$, so
\begin{align}
    &\mathbb{E}[\delta p_i^R] = \begin{bmatrix}
        \rho_i\sin\theta_i\cos\phi_i(e^{-\frac{\epsilon_\theta^2+\epsilon_\phi^2}{2}}-1)\\
        \rho_i\sin\theta_i\sin\phi_i(e^{-\frac{\epsilon_\theta^2+\epsilon_\phi^2}{2}}-1)\\
        \rho_i\cos\theta_i(e^{-\frac{\epsilon_\theta^2}{2}}-1)
    \end{bmatrix}\\
\end{align}
it can be concluded that its expectation is not equal to zero if $\epsilon_\theta>0$ or $\epsilon_\phi>0$.

The error covariance propagation is derived using the Jacobian matrix approximation. The Jacobian matrix $\mathbf{J}_i$ for error propagation from spherical to Cartesian coordinates is:
\begin{equation}
    \mathbf{J}_i = \begin{bmatrix}
        \sin\theta_i\cos\phi_i & \rho_i\cos\theta_i\cos\phi_i & -\rho_i\sin\theta_i\sin\phi_i \\
        \sin\theta_i\sin\phi_i & \rho_i\cos\theta_i\sin\phi_i & \rho_i\sin\theta_i\cos\phi_i \\
        \cos\theta_i & -\rho_i\sin\theta_i & 0
    \end{bmatrix}
\end{equation}
The covariance matrix in Cartesian coordinates is:
\[
\boldsymbol{\Sigma}_{\text{C}} = \mathbf{J} \boldsymbol{\Sigma}_{\text{S}} \mathbf{J}^\top
\]
where $\boldsymbol{\Sigma}_{\text{C}}$ represents the first-order error propagation.

For the PnP problem, the projection equation with noise propagation is:

\begin{equation}
    s_i\begin{bmatrix}
        \Tilde{q}_i \\ 1
    \end{bmatrix} = K[R|t]\left(\begin{bmatrix}
        p_i \\ 1
    \end{bmatrix} + \begin{bmatrix}
        \delta p_i \\ 0
    \end{bmatrix}\right)
\end{equation}

where $\delta p_i \sim \mathcal{N}(\mathbb{E}[\delta p_i], \boldsymbol{\Sigma}_C)$ represents the propagated noise in Cartesian coordinates. The maximum likelihood estimate of camera pose $(R,t)$ minimizes the Mahalanobis distance $r_M$:

\begin{equation}
    \arg\min_{R,t} \sum_{i=1}^n \left\| \Tilde{p_i} - R^{-1}(s_i K^{-1}[q_i^T, 1]^T-t)-\mathbb{E}[\delta p_i] \right\|_{\boldsymbol{\Sigma}_C^{-1}}^2
\end{equation}

where $s_i=\Tilde{p}_i^{(3)}$ . Then, the LM optimization algorithm is used to optimize the function and solve for the optimal R and t.

\subsection{4D Radar-Camera Calibration System Design}


This section presents a millimeter-wave radar-camera extrinsic calibration system aimed at developing an accurate, reliable, and low-cost solution. A key innovation is the replacement of conventional multi-reflector simultaneous recognition with temporal accumulation of single-reflector observations. This approach simplifies hardware requirements and enhances system stability.

The calibration workflow supports three operational modes through dedicated processing pipelines, improving usability. As illustrated in Fig. 2, the system architecture comprises two core modules: 1) a feature pairs detector (integrating 2D image and 3D radar point detection), and 2) the 3DUPnP solver. This integrated design balances calibration precision, implementation complexity, and operational convenience.

\subsubsection{Feature Pairs Detector} 
The detection system comprises 2D image-based detection and 3D spatial point detection modules. The implementation details are as follows:

The 2D detection module was trained on 1,191 manually annotated images with significant inter-frame variations, including indoor/outdoor scenes at 2-15m distances under varying illumination. Training followed the YOLOv8 framework\cite{b22}, which enhances robustness and incorporates the known geometric structure of targets (Fig. 4) to compute image coordinates of reflection hotspots - corresponding to 3D points with peak intensity.

The 3D detector exploits millimeter-wave radar's (MMW) active emission principle and corner reflector geometry to generate high-intensity spatial points. After denoising and range truncation \cite{b19}, points within intensity thresholds are selected. We extract the intensity-weighted centroid corresponding to the inner corner.

We designed two time synchronizers. The first one achieves accurate radar point-to-image feature registration by applying B-spline interpolation to M data pairs within N sliding windows. This method ensures spatiotemporal correspondence while effectively correcting timestamp misalignment, and it is motion-agnostic. The second one collects M data pairs in stationary state over N time intervals, first identifies N sets of distributions, and then feeds them into the backend optimizer. Experiments demonstrate that using the detection results from the latter for registration can significantly suppress error propagation while reducing noise interference.

\subsubsection{3DUPnP solver with RANSAC} 
 The solver architecture comprises three core components:
\begin{itemize}
    \item Initialization Solver: Provides initial estimates of rotation matrix R and translation vector t through established linear solvers (e.g., EPnP or CPnP).
    \item Optimization Module: Refines initial estimates by minimizing the error function defined in Section 3.1 through iterative nonlinear optimization.
    \item RANSAC Inlier Filter: Implements outlier rejection for measurement point pairs using RANSAC-based consensus filtering to enhance robustness.
\end{itemize}
The solver pseudocode is structured as algorithm1.

\begin{algorithm}[t]
\caption{3DUPnP Solver with RANSAC}
\label{alg:3DUPnP}
\SetAlgoLined
\KwIn{Correspondences $\{(P_i, q_i)\}_{i=1}^n$, confidence $p$, sample size $s=4$, min inlier ratio $\rho_{\min}$}
\KwOut{Optimal $R^*, t^*$, inlier indices $\mathcal{I}^*$}
Initialize $N \gets \infty$, $iter \gets 0$, $\rho^* \gets 0$\;
\While{$iter < N$ \textbf{and} $\rho^* < \rho_{\min}$}{
    Randomly sample $s$ correspondences $\mathcal{S}$\;
    $R_0, t_0 \gets \text{EPnP}(\mathcal{S})$\; 
    $R, t \gets \text{LM-Optimize}(R_0, t_0, \text{Eq.}(6))$\;
    Compute residuals $\{r_i\}$ via Eq.(6) for all points\;
    $\mathcal{I} \gets \{i | r_i < \tau\}$ 
    $\rho \gets |\mathcal{I}| / n$ 
    \If{$\rho > \rho^*$}{
        $\rho^* \gets \rho$\;
        $R^* \gets R$, $t^* \gets t$\;
        $\mathcal{I}^* \gets \mathcal{I}$\;
        $N \gets \min\left(N, \frac{\log(1-p)}{\log(1 - (\rho^*)^s)}\right)$ 
    }
    $iter \gets iter + 1$\;
}
\Return $R^*, t^*, \mathcal{I}^*$\;
\end{algorithm}
\section{EXPERIMENT}
This section presents the experimental validation framework, comprising simulation-based verification and real-world platform experiments. The simulation studies aim to validate the consistency of the proposed 3DUPnP algorithm. The physical experiments evaluate the superiority of our method against traditional and state-of-the-art approaches.
\subsection{Simulation}
\begin{center}\vspace{1cm}
\includegraphics[width=0.85\linewidth]{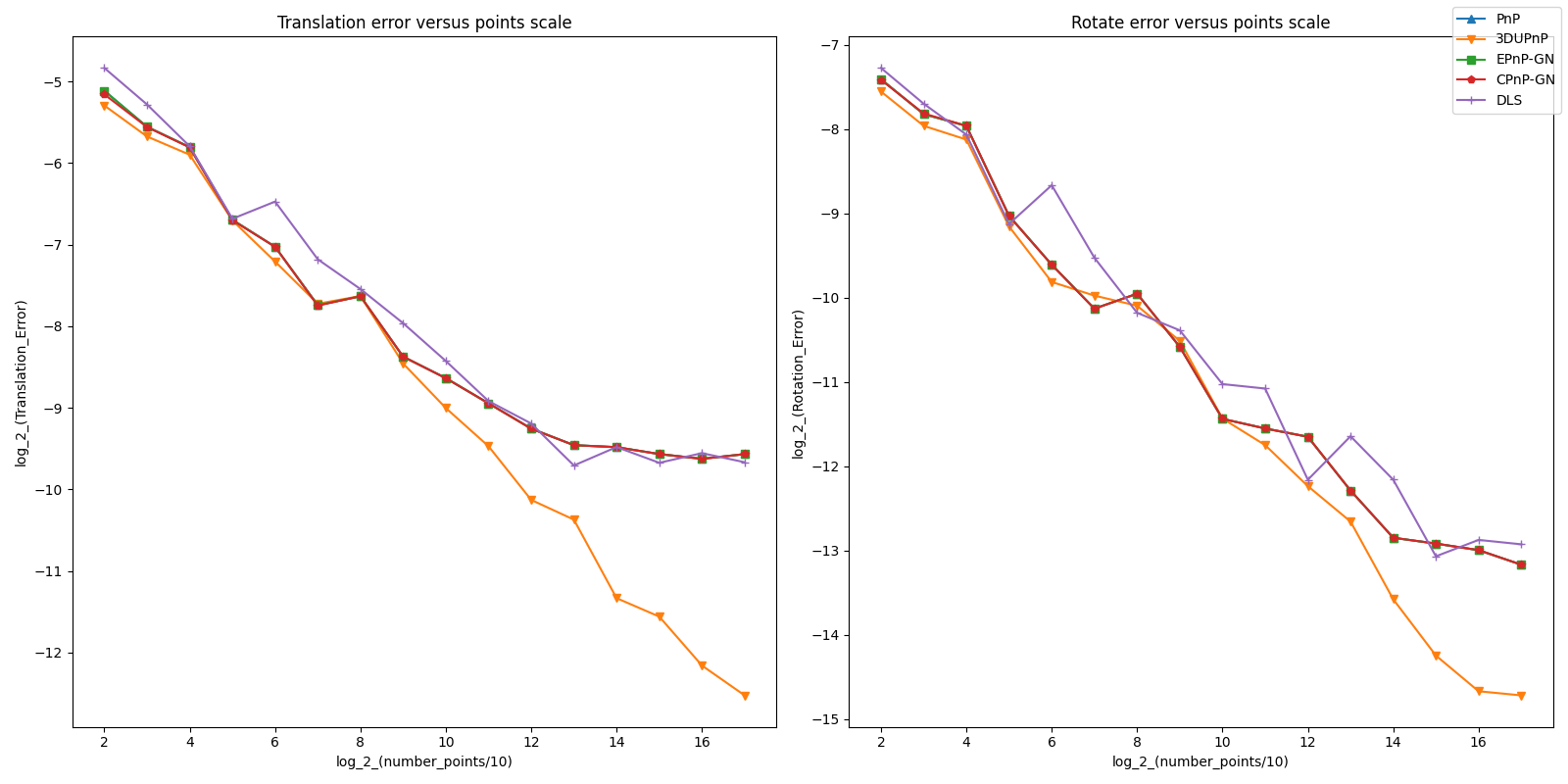}
\captionof{figure}{Consistency validation result}
\label{consistency_validation}
\end{center}\vspace{1cm}
The simulation, conducted in Python, generates N random points in space, adding observation noise with $\sigma_\rho=0.02m,\sigma_\theta=\sigma_\phi=0.005rad$ in spherical coordinates(according to the parameters zf provided). A 30-degree orientation difference between the camera and radar is included in the true extrinsic parameters. Consistency verification begins with N=10 points, doubling each time, comparing the proposed method and other methods against ground truth. Errors are calculated as the norm of Euler angles for rotation and the translation vector. The proposed 3DUPnP method outperforms other methods in terms of translation consistency and rotation accuracy as Fig.3 shows.

\subsection{Real World Experiment}
The calibration ground truth is derived from CAD-designed 3D-printed mounting fixtures and sensor-specific physical models. Subsequent manual refinement of extrinsic parameters was performed using OpenCalib\cite{b4}, with final validation conducted in outdoor long-range scenarios involving handheld corner reflectors, confirming the reference accuracy under practical operating conditions.
\begin{figure*}[thpb]
\centering
\includegraphics[width=0.75\linewidth]{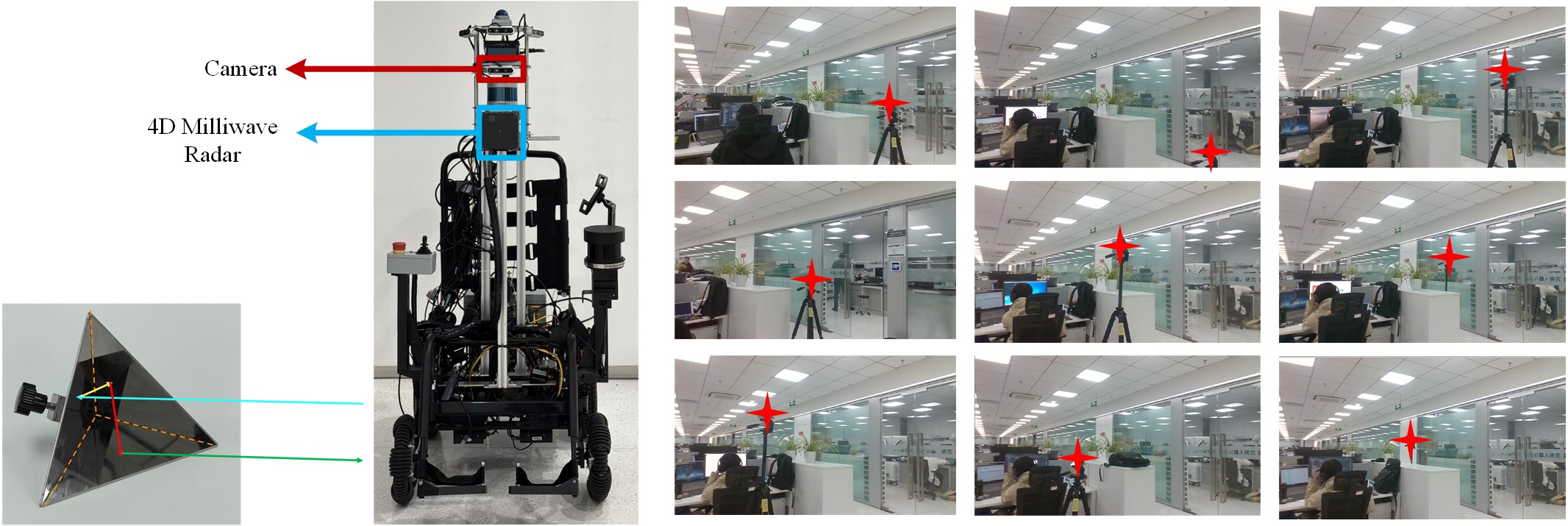}
\caption{Corner reflector provide high-intensity feature, the sensors installed on our smart wheelchair, and physical experiment conducted at the office. *The red star marks the location of the corner reflector.}
\label{fig:scene_robot_reflector}
\end{figure*}

\begin{figure*}[thpb]
\centering
\includegraphics[width=0.85\linewidth]{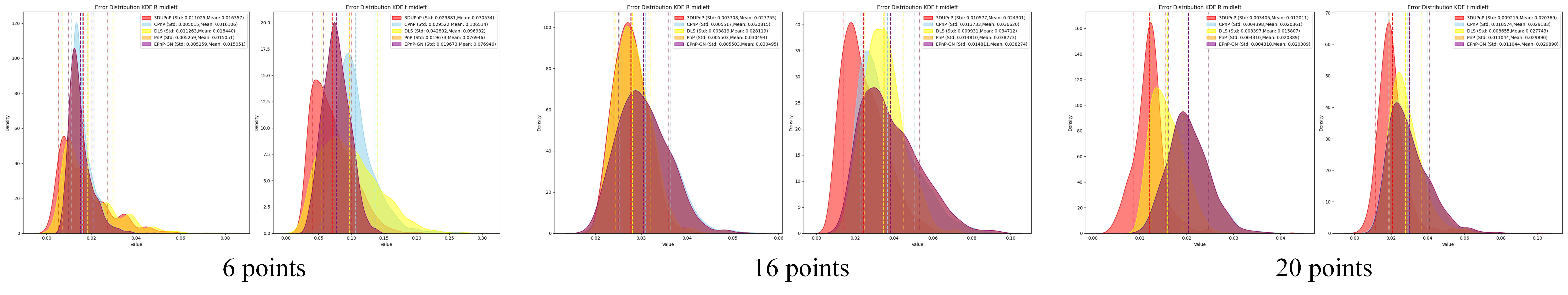}
\caption{Kernel density estimation (KDE) plots of calibration errors corresponding to 6, 16, and 20 calibration points.}
\label{fig:KDE}
\end{figure*}
Under the experimental configuration illustrated in Fig.4, our system was evaluated with varying numbers of calibration points (6, 16, and 20). For each point, 100 measurements were randomly subsampled, followed by extrinsic parameter estimation and comparison against ground truth. This process was repeated 1,000 times to generate kernel density estimation (KDE) plots of error distributions as shown in Fig. 5, and subsequently compute the comparison of rotation errors and translation errors for different PnP algorithms as presented in TABLE 1, enabling statistically robust analysis of calibration precision under stochastic measurement variations.
\begin{table}[h]
\caption{Mean Errors of Different PnP Algorithms}

\label{result}
\begin{center}
\begin{tabular}{|c|c|c|c|c|}
\hline
 & & 6 Points & 16 Points & 20 Points \\
\hline
\multirow{2}{*}{3DUPnP}
& $e_{rot}(rad)$ &0.016357 &\textbf{0.027755} &\textbf{0.012011} \\
& $e_{tran}(m)$ &\textbf{0.070534} &\textbf{0.024301} & \textbf{0.020769}\\
\hline
\multirow{2}{*}{CPnP-GN}
& $e_{rot}$ &0.016106 &0.031815 &0.020361 \\
& $e_{tran}$ & 0.106514&0.036620 & 0.029183\\
\hline
\multirow{2}{*}{DLS}
& $e_{rot}$ &0.01844 &0.028119 &0.015807 \\
& $e_{tran}$ &0.096932 & 0.034712& 0.027743\\
\hline
\multirow{2}{*}{PnP}
& $e_{rot}$ &\textbf{0.015051} & 0.030494& 0.020389\\
& $e_{tran}$ &0.076946 & 0.038273& 0.029890\\
\hline
\multirow{2}{*}{EPnP-GN}
& $e_{rot}$ &\textbf{0.015051} &0.030495 & 0.020389\\
& $e_{tran}$ &0.076946 &0.038274 &0.029890 \\
\hline
\end{tabular}
\end{center}
\end{table}
The analysis reveals that our method achieves significantly superior accuracy compared to existing approaches under dense calibration point configurations. While sparse point conditions exhibit multimodal error distributions, the primary mode still surpasses competing algorithms in precision. The attached video recordings also demonstrate high consistency in the calibration results.

\section{CONCLUSIONS}
We present an accurate, reliable, low-cost calibration framework for 4D radar-camera systems, advancing the integration of cost-effective, environment-resilient MMW radars into perception pipelines. As reliable calibration underpins sensor fusion, future work will focus on standardization and active perception-integrated calibration to enhance robustness.











\begin{thebibliography}{99}

\bibitem{b1} K. Harlow, H. Jang, T. D. Barfoot, A. Kim and C. Heckman, "A New Wave in Robotics: Survey on Recent MmWave Radar Applications in Robotics," in IEEE Transactions on Robotics, vol. 40, pp. 4544-4560, 2024, doi: 10.1109/TRO.2024.3463504.
\bibitem{b2} Zhou, Y.; Dong, Y.; Hou, F.; Wu, J. Review on Millimeter-Wave Radar and Camera Fusion Technology. Sustainability 2022, 14, 5114. https://doi.org/10.3390/su14095114.
\bibitem{b3} S. Urban, J. Leitloff, and S. Hinz, “Mlpnp-a real-time maximum likelihood solution to the perspective-n-point problem,” arXiv preprint arXiv:1607.08112, 2016.
\bibitem{b4} G. Yan, L. Zhuochun, C. Wang, C. Shi, P. Wei, X. Cai, T. Ma, Z. Liu, Z. Zhong, Y. Liu, M. Zhao, Z. Ma, and Y. Li, "OpenCalib: A Multi-sensor Calibration Toolbox for Autonomous Driving," arXiv, 2022. [Online]. Available: https://arxiv.org/abs/2205.14087.
\bibitem{b5} J. Peršić, I. Marković, and I. Petrović, "Extrinsic 6DoF calibration of a radar–LiDAR–camera system enhanced by radar cross section estimates evaluation," Robotics and Autonomous Systems, vol. 114, pp. 217-230, 2019. 
\bibitem{b6} J. Domhof, J. F. P. Kooij and D. M. Gavrila, "A Joint Extrinsic Calibration Tool for Radar, Camera and Lidar," in IEEE Transactions on Intelligent Vehicles, vol. 6, no. 3, pp. 571-582, Sept. 2021, doi: 10.1109/TIV.2021.3065208.
\bibitem{b7} J. Domhof, J. F. P. Kooij and D. M. Gavrila, "An Extrinsic Calibration Tool for Radar, Camera and Lidar," 2019 International Conference on Robotics and Automation (ICRA), Montreal, QC, Canada, 2019, pp. 8107-8113, doi: 10.1109/ICRA.2019.8794186.
\bibitem{b8} J. Zhang et al., "LB-R2R-Calib: Accurate and Robust Extrinsic Calibration of Multiple Long Baseline 4D Imaging Radars for V2X," 2024 IEEE International Conference on Robotics and Automation (ICRA), Yokohama, Japan, 2024, pp. 16729-16735, doi: 10.1109/ICRA57147.2024.10611470.
\bibitem{b9} Y. Wang, W. Zhao, C. Cao, T. Deng, J. Wang, and W. Chen, “SFPNet: Sparse Focal Point Network for Semantic Segmentation on General LiDAR Point Clouds,” in Computer Vision – ECCV 2024: 18th European Conference on Computer Vision, Proceedings, 2025, vol. 14851, pp. 403–421. doi: 10.1007/978-3-031-72652-1-24.
\bibitem{b10} E. Wise, Q. Cheng and J. Kelly, "Spatiotemporal Calibration of 3-D Millimetre-Wavelength Radar-Camera Pairs," in IEEE Transactions on Robotics, vol. 39, no. 6, pp. 4552-4566, Dec. 2023, doi: 10.1109/TRO.2023.3311680.
\bibitem{b11} E. Wise, J. Peršić, C. Grebe, I. Petrović and J. Kelly, "A Continuous-Time Approach for 3D Radar-to-Camera Extrinsic Calibration," 2021 IEEE International Conference on Robotics and Automation (ICRA), Xi'an, China, 2021, pp. 13164-13170, doi: 10.1109/ICRA48506.2021.9561938.
\bibitem{b12} J. Peršić, L. Petrović, I. Marković and I. Petrović, "Spatiotemporal Multisensor Calibration via Gaussian Processes Moving Target Tracking," in IEEE Transactions on Robotics, vol. 37, no. 5, pp. 1401-1415, Oct. 2021, doi: 10.1109/TRO.2021.3061364.
\bibitem{b13} C. Schöller, M. Schnettler, A. Krämmer, G. Hinz, M. Bakovic, M. Güzet, and A. Knoll, "Targetless rotational auto-calibration of radar and camera for intelligent transportation systems," arXiv, 2019. [Online].
\bibitem{b14} A. Vakhitov, L. F. Colomina, A. Agudo, and F. Moreno-Noguer, "Uncertainty-aware camera pose estimation from points and lines," arXiv, 2021. [Online].
\bibitem{b15} J. A. Hesch and S. I. Roumeliotis, "A Direct Least-Squares (DLS) method for PnP," 2011 International Conference on Computer Vision, Barcelona, Spain, 2011, pp. 383-390, doi: 10.1109/ICCV.2011.6126266.
\bibitem{b16} S. Li, C. Xu and M. Xie, "A Robust O(n) Solution to the Perspective-n-Point Problem," in IEEE Transactions on Pattern Analysis and Machine Intelligence, vol. 34, no. 7, pp. 1444-1450, July 2012, doi: 10.1109/TPAMI.2012.41.
\bibitem{b17} G. Zeng, S. Chen, B. Mu, G. Shi and J. Wu, "CPnP: Consistent Pose Estimator for Perspective-n-Point Problem with Bias Elimination," 2023 IEEE International Conference on Robotics and Automation (ICRA), London, United Kingdom, 2023, pp. 1940-1946, doi: 10.1109/ICRA48891.2023.10160942.
\bibitem{b18} V. Lepetit, F. Moreno-Noguer, and P. Fua, “Epnp: An accurate o (n)solution to the pnp problem,” International Journal of Computer Vision,vol. 81, no. 2, pp. 155–166, 2009.
\bibitem{b19} X. Li, H. Zhang and W. Chen, "4D Radar-Based Pose Graph SLAM With Ego-Velocity Pre-Integration Factor," in IEEE Robotics and Automation Letters, vol. 8, no. 8, pp. 5124-5131, Aug. 2023, doi: 10.1109/LRA.2023.3292574. 
\bibitem{b20} Xiao-Shan Gao, Xiao-Rong Hou, Jianliang Tang and Hang-Fei Cheng, "Complete solution classification for the perspective-three-point problem," in IEEE Transactions on Pattern Analysis and Machine Intelligence, vol. 25, no. 8, pp. 930-943, Aug. 2003, doi: 10.1109/TPAMI.2003.1217599. 
\bibitem{b21} Peršić J, Petrović L, Marković I, et al. Online multi-sensor calibration based on moving object tracking[J]. Advanced Robotics, 2021, 35(3-4): 130-140.
\bibitem{b22} R. Varghese and S. M., "YOLOv8: A Novel Object Detection Algorithm with Enhanced Performance and Robustness," 2024 International Conference on Advances in Data Engineering and Intelligent Computing Systems (ADICS), Chennai, India, 2024, pp. 1-6, doi: 10.1109/ADICS58448.2024.10533619.
\bibitem{b23} Adrian Penate-Sanchez, Juan Andrade-Cetto, and Francesc Moreno-Noguer. Exhaustive linearization for robust camera pose and focal length estimation. Pattern Analysis and Machine Intelligence, IEEE Transactions on, 35(10):2387–2400, 2013.
\bibitem{b24} T. Deng et al., "MNE-SLAM: Multi-agent neural SLAM for mobile robots," in Proc. IEEE/CVF Conf. Comput. Vis. Pattern Recognit. (CVPR), 2025, pp. 1485-1494.



\end{thebibliography}
\end{document}